%% file: Template.tex
\documentclass{article}
\usepackage{spconf,amsmath,graphicx,hyperref}
\usepackage{mathrsfs}
\usepackage[dvipsnames]{xcolor}
\usepackage[normalem]{ulem}
\useunder{\uline}{\ul}{}
\usepackage{amsmath,amsfonts}
\usepackage{amssymb}
\usepackage{algorithmic}

\usepackage{array}

\usepackage{textcomp}
\usepackage{stfloats}
\usepackage{url}
\usepackage{verbatim}

\usepackage{cite}
\usepackage{longtable}
\usepackage{booktabs}
\usepackage{multirow}

\usepackage{graphicx}

\usepackage[table,xcdraw]{xcolor}
\usepackage{colortbl}
\usepackage[textsize=tiny]{todonotes}
\usepackage{bm}
\usepackage{fix-cm}

\usepackage{pifont}
\usepackage{tikz}

\usepackage[caption=false,font=scriptsize,labelfont=rm,textfont=rm]{subfig}

\usetikzlibrary{shadows.blur}
\usepackage[most]{tcolorbox}
\tcbuselibrary{skins}

\tcbuselibrary{skins,breakable}

\usepackage{hyperref}

\usepackage{float}
\usepackage[ruled,linesnumbered,vlined,noend]{algorithm2e}

\hypersetup{
    colorlinks=true,
    citecolor=brown, 
    linkcolor=brown,
    urlcolor=brown
}

\definecolor{lightblue}{RGB}{204, 236, 230}

\definecolor{lightteal}{RGB}{204, 236, 230}

\definecolor{softteal}{RGB}{204, 236, 230}
\definecolor{mediumteal}{RGB}{204, 236, 230}

\SetKwBlock{Begin}{}{}
\SetKwIF{If}{ElseIf}{Else}{if}{then}{else if}{else}{}
\SetKwFor{For}{for}{do}{}

\SetKwInput{KwIn}{\textbf{Input}}
\SetKwInput{KwOut}{\textbf{Output}}

\newcounter{myexample}

\newcounter{myquestion}

\tcbset{
  enhanced,                       
  colframe=brown,        
  colback=brown!6!white,          
  boxrule=0.8mm,                  
  arc=2mm,                        
  drop shadow=brown!70!black,    
  left=2mm, right=2mm, top=1mm, bottom=1mm, 
  before skip=12pt,               
  after skip=12pt                
}

\title{Style Attack Disguise: When Fonts Become a Camouflage for Adversarial Intent}
\name{Yangshijie Zhang$^{\dagger 1}$ \, Xinda Wang$^{\dagger 2}$ \, Jialin Liu$^{2}$ \, Wenqiang Wang$^{3}$ \, Zhicong Ma$^{1}$ \, Xingxing Jia$^{\star 1}$\thanks{$^{\dagger}$ These authors contributed equally. $^{\star}$ Corresponding author: jiaxx@lzu.edu.cn}
\thanks{© 2026 IEEE. Personal use of this material is permitted. Permission from IEEE must be obtained for all other uses, in any current or future media, including reprinting/republishing this material for advertising or promotional purposes, creating new collective works, for resale or redistribution to servers or lists, or reuse of any copyrighted component of this work in other works.}}
\address{$^{1}$Lanzhou University \qquad $^{2}$Peking University \qquad $^{3}$Sun Yat-sen University}

\begin{document}
\ninept
\maketitle
\begin{abstract}
With social media growth, users employ stylistic fonts and font-like emoji to express individuality, creating visually appealing text that remains human-readable. However, these fonts introduce hidden vulnerabilities in NLP models: while humans easily read stylistic text, models process these characters as distinct tokens, causing interference. We identify this human-model perception gap and propose a style-based attack, Style Attack Disguise (SAD). We design two sizes: light for query efficiency and strong for superior attack performance. Experiments on sentiment classification and machine translation across traditional models, LLMs, and commercial services demonstrate SAD's strong attack performance. We also show SAD's potential threats to multimodal tasks including text-to-image and text-to-speech generation.
\end{abstract}
\begin{keywords}
Adversarial attack, style-based attack, stylistic fonts, font-like emoji
\end{keywords}
\section{Introduction}
\label{sec:intro}

Social media users decorate text with special fonts like mathematical alphabets ($\mathcal{A}$), regional indicator symbols (\raisebox{-0.6ex}{\hspace{0.8pt}\includegraphics[height=0.35cm,width=0.35cm]{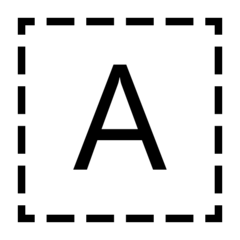}}\hspace{1pt}), and squared letters (\raisebox{-0.8ex}{\hspace{0.8pt}\includegraphics[height=0.51cm,width=0.51cm]{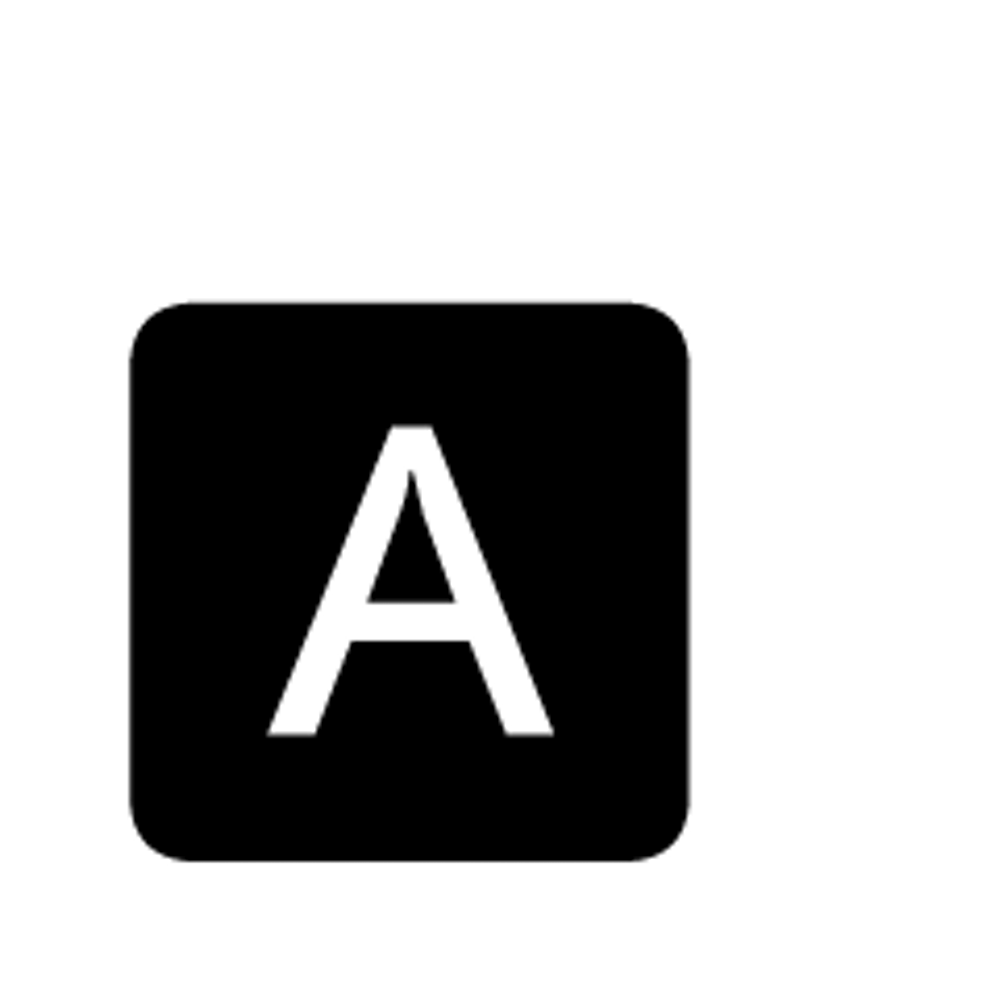}})~\cite{aliprand2000unicode}. These fonts remain user-friendly to read while expressing individuality.

\begin{figure}[t]
\centering
\includegraphics[width=0.71\columnwidth]{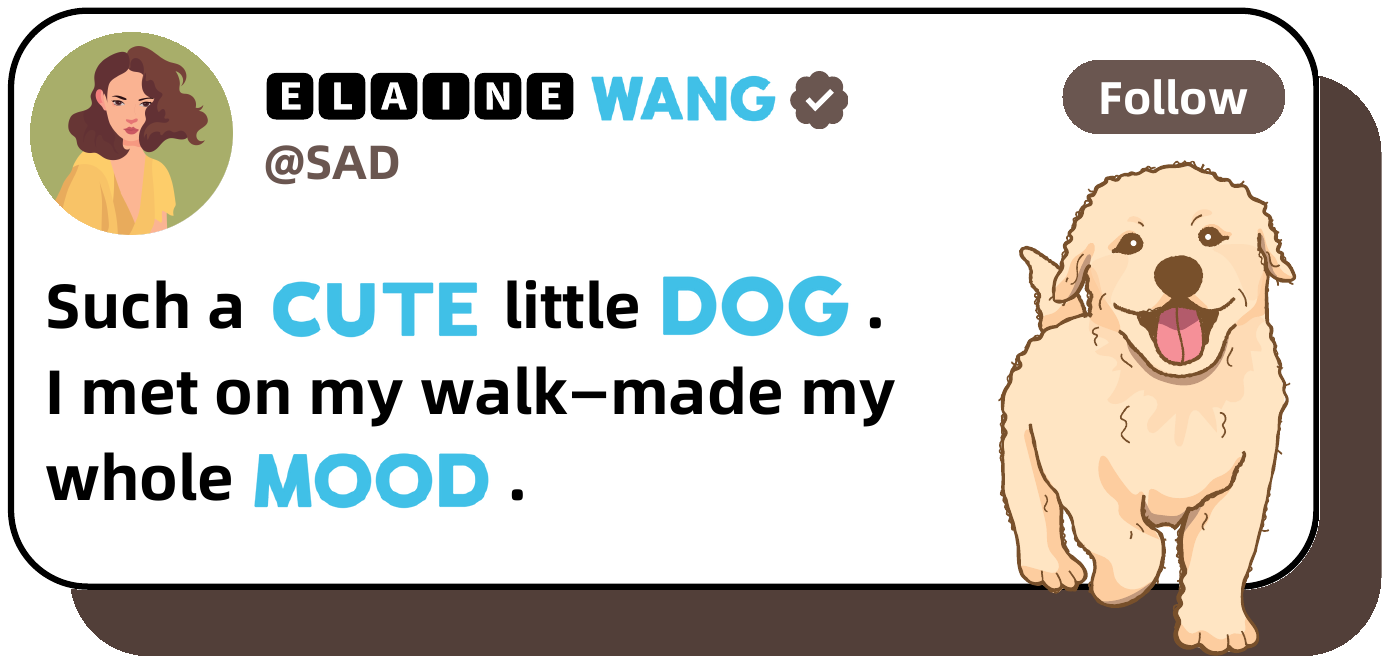}
\caption{On Twitter (X), a user decorate display names with stylized fonts and use special font-like emoji in tweets to highlight key words and express individuality.}
\label{fig:twitter}
\end{figure}

Regional indicator symbols exemplify this trend. While most users know flag emojis like \raisebox{-0.6ex}{\hspace{0.8pt}\includegraphics[height=0.35cm,width=0.35cm]{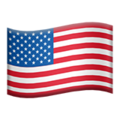}}\hspace{1pt}, few realize these are composed of two regional indicator symbols (\raisebox{-0.6ex}{\hspace{0.8pt}\includegraphics[height=0.35cm,width=0.35cm]{img/us.png}}\hspace{1pt} = \raisebox{-0.6ex}{\hspace{0.8pt}\includegraphics[height=0.35cm,width=0.35cm]{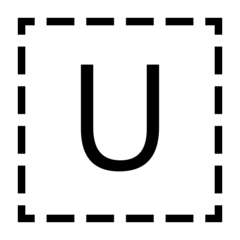}}\hspace{1pt} + \raisebox{-0.6ex}{\hspace{0.8pt}\includegraphics[height=0.35cm,width=0.35cm]{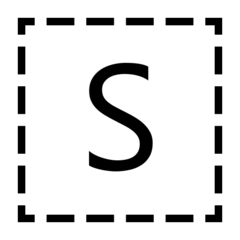}}\hspace{1pt})~\cite{kariryaa2022role}. When these symbols appear individually, many platforms render them as stylized blue letters (\raisebox{-0.3ex}{\hspace{0.8pt}\includegraphics[height=0.28cm,width=0.28cm]{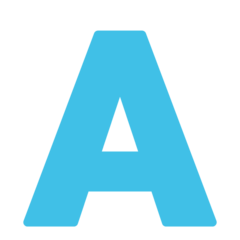}}\hspace{1pt}). Users can spell words using these \textit{font-like emoji}, as shown in Figure~\ref{fig:twitter}, to highlight key words and express individuality in their posts~\cite{ge2021communicative}.
This creates an exploitable vulnerability: humans perceive identical meaning across different font styles and read styled text normally, while models trained on standard text process these characters as distinct tokens~\cite{garcia2022out}. \textit{This human-model perception gap leads to inconsistent behaviors when models encounter styled text.}
However, prior work has rarely focused on the phenomenon of widespread stylistic font usage and its security risks~\cite{wang2022measure,formentoconfidence}.

Consider this message: ``How many \raisebox{-0.3ex}{\hspace{0.8pt}\includegraphics[height=0.28cm,width=0.28cm]{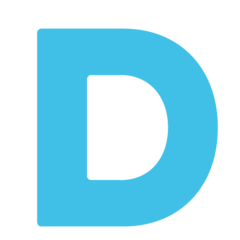}}\raisebox{-0.3ex}{\hspace{0.8pt}\includegraphics[height=0.28cm,width=0.28cm]{img/android-a.png}}\raisebox{-0.3ex}{\hspace{0.8pt}\includegraphics[height=0.28cm,width=0.28cm]{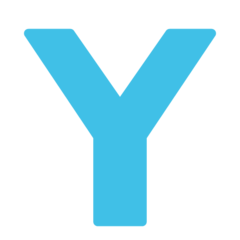}}\raisebox{-0.3ex}{\hspace{0.8pt}\includegraphics[height=0.28cm,width=0.28cm]{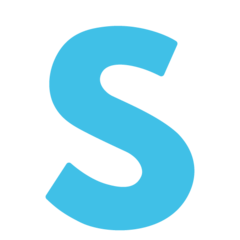}} are there in a \raisebox{-0.3ex}{\hspace{0.8pt}\includegraphics[height=0.28cm,width=0.28cm]{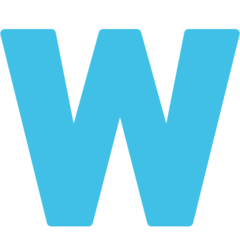}}\raisebox{-0.3ex}{\hspace{0.8pt}\includegraphics[height=0.28cm,width=0.28cm]{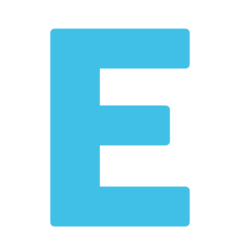}}\raisebox{-0.3ex}{\hspace{0.8pt}\includegraphics[height=0.28cm,width=0.28cm]{img/e-regional.png}}\raisebox{-0.3ex}{\hspace{0.8pt}\includegraphics[height=0.28cm,width=0.28cm]{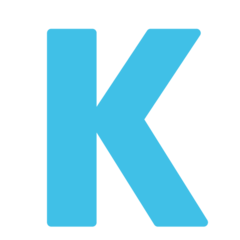}}?" Humans read this normally as ``How many DAYS are there in a WEEK?" while models struggle with these stylistic fonts, leading to misinterpretation. This example illustrates a broader phenomenon: whether using regional indicator symbols, mathematical alphabets ($\mathbb{DAYS}$), or circled letters (\raisebox{-1.35ex}{\hspace{0.8pt}\includegraphics[height=0.51cm,width=1.2cm]{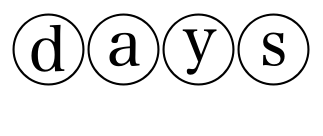}}), all create the same human-model perception gap.

Motivated by this observation, we propose \textbf{\textcolor{brown}{S}}tyle \textbf{\textcolor{brown}{A}}ttack \textbf{\textcolor{brown}{D}}isguise (\textcolor{brown}{SAD}), an attack that uses stylistic fonts to fool models while remaining human-readable.
SAD offers two sizes that provide flexible trade-offs between attack intensity and query efficiency in our experiments. Additionally, SAD can readily incorporate new stylistic fonts as they emerge, providing a plug-and-play framework.
Our contributions are:

\begin{tcolorbox}
\ding{182} We introduce a style-level adversarial attack that exploits stylistic fonts while maintaining visual readability.

\ding{183} We develop a hybrid word ranking method that balances semantic importance and tokenization instability for optimal target selection.

\ding{184} We demonstrate effectiveness across WordPiece, BPE, and large language model architectures.
\end{tcolorbox}

\section{Related Work}

Text adversarial attacks can be categorized by their operational levels. \textbf{Character-level attacks} target individual characters through typos, swaps, or insertions~\cite{belinkov2018synthetic,gao2018deepwordbug,li2019textbugger}, but often produce visually detectable changes. \textbf{Word-level attacks} manipulate tokens through synonym replacement~\cite{jin2020bert,ren2019generating}, reordering~\cite{moradi2021evaluating}, or deletion~\cite{xie2022word}, with frameworks like TextAttack~\cite{morris2020textattack} providing systematic evaluation. \textbf{Sentence-level attacks} insert misleading content or generate adversarial sequences~\cite{lin2021using,huang2021generating} while maintaining semantic consistency.

\section{Method}

\begin{figure*}[t]
\centering
\includegraphics[width=0.98\textwidth]{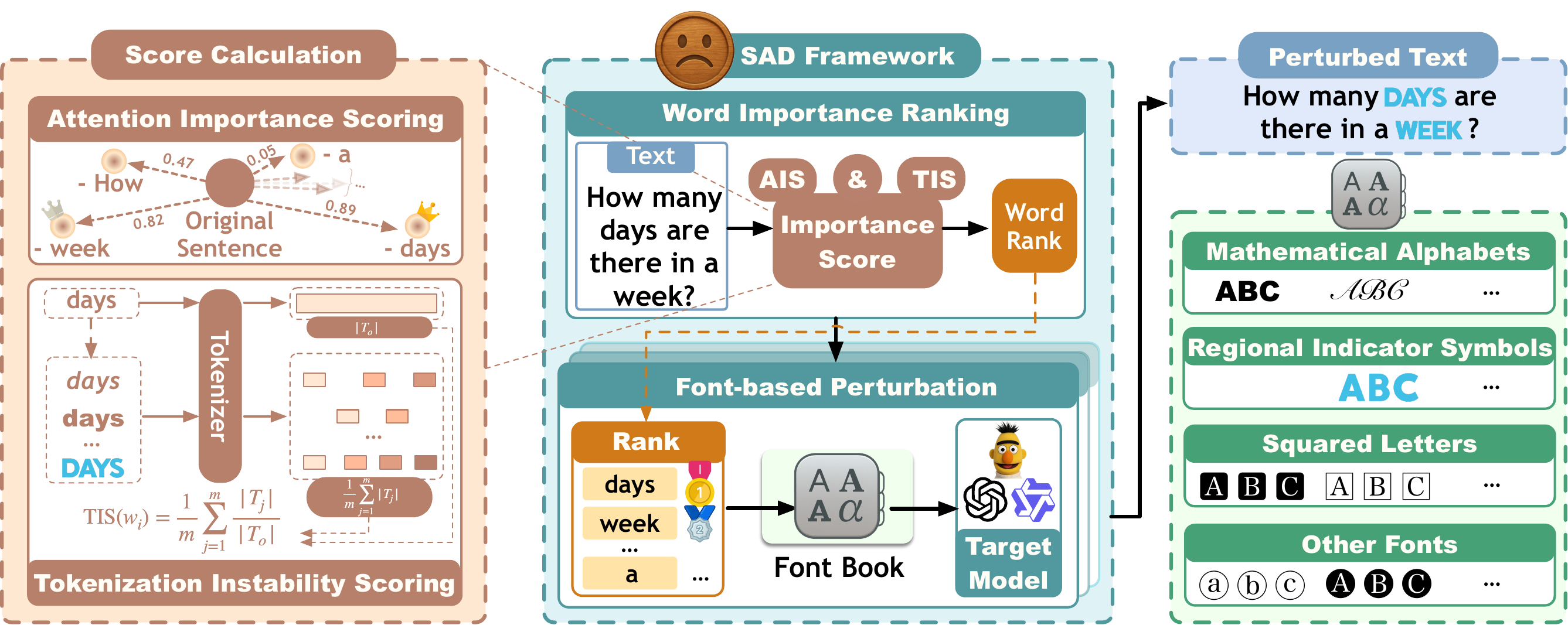}
\caption{SAD first ranks words by semantic importance (AIS) and tokenization instability (TIS) to establish attack priority, then iteratively apply font-based perturbations to candidate words until successful. Here, we replace ``DAYS" with ``\raisebox{-0.3ex}{\hspace{0.8pt}\includegraphics[height=0.28cm,width=0.28cm]{img/d-regional.png}}\raisebox{-0.3ex}{\hspace{0.8pt}\includegraphics[height=0.28cm,width=0.28cm]{img/android-a.png}}\raisebox{-0.3ex}{\hspace{0.8pt}\includegraphics[height=0.28cm,width=0.28cm]{img/y-regional.png}}\raisebox{-0.3ex}{\hspace{0.8pt}\includegraphics[height=0.28cm,width=0.28cm]{img/s-regional.png}}'' and ``WEEK" with ``\raisebox{-0.3ex}{\hspace{0.8pt}\includegraphics[height=0.28cm,width=0.28cm]{img/w-regional.png}}\raisebox{-0.3ex}{\hspace{0.8pt}\includegraphics[height=0.28cm,width=0.28cm]{img/e-regional.png}}\raisebox{-0.3ex}{\hspace{0.8pt}\includegraphics[height=0.28cm,width=0.28cm]{img/e-regional.png}}\raisebox{-0.3ex}{\hspace{0.8pt}\includegraphics[height=0.28cm,width=0.28cm]{img/k-regional.png}}'', keeping text readable for humans but confusing for models.}
\label{fig:overview}
\end{figure*}

Figure~\ref{fig:overview} shows our SAD framework, which exploits stylistic fonts through font-based perturbation and word importance ranking.

\subsection{Font-based Perturbation}

Formally, let $\mathcal{X}$ be the input text space and $\mathcal{Y}$ be the output label space. The target model $f_{\text{t}}: \mathcal{X} \to \mathcal{Y}$ maps input text to predicted labels. For text $x \in \mathcal{X}$, we have $y = f_{\text{t}}(x)$.

We propose two attack modes: $\text{SAD}_{\text{light}}$ and $\text{SAD}_{\text{strong}}$. $\text{SAD}_{\text{light}}$ employs a query-efficient approach with maximum $T$ queries, while $\text{SAD}_{\text{strong}}$ performs comprehensive substitution in a single query.
Let $\mathcal{C}$ be the set of standard characters and $\mathcal{S}$ be the set of their stylistic font forms. We define the font substitution space as:
\begin{equation}
\label{eq:substitution_space}
\mathcal{S} = \mathcal{M} \cup \mathcal{R} \cup \mathcal{O} \cup \mathcal{Q} \cup \mathcal{V},
\end{equation}
where $\mathcal{M}$, $\mathcal{R}$, $\mathcal{O}$, $\mathcal{Q}$, $\mathcal{V}$ represent mathematical alphabets, regional indicator symbols, circled letters, squared letters, and other stylistic fonts, respectively. Each character $c \in \mathcal{C}$ has corresponding stylistic forms $\{s_1, s_2, \ldots, s_k\} \subset \mathcal{S}$ that look the same but are actually distinct characters.
Given text $x$ with words $\{w_1, w_2, \ldots, w_n\}$ ranked by importance, we define the substitution function $\phi: \mathcal{C} \to \mathcal{S}$ that maps standard characters to their stylistic fonts. For a subset of words $W \subseteq \{w_1, w_2, \ldots, w_n\}$, the perturbed text is:
\begin{equation}
\label{eq:perturbation}
\tilde{x} = f_{\text{sub}}(x, W, \phi),
\end{equation}
where $f_{\text{sub}}$ performs character substitution and characters in words $W$ are replaced using function $\phi$.

For $\text{SAD}_{\text{light}}$, we gradually increase the number of perturbed words from a single word to larger subsets until attack success or query budget $T$ is reached. For $\text{SAD}_{\text{strong}}$, all words are perturbed simultaneously: $W = \{w_1, w_2, \ldots, w_n\}$.
Our objective is finding $\tilde{x}$ that satisfies $f_{\text{t}}(\tilde{x}) \neq f_{\text{t}}(x)$ while maintaining visual similarity between $x$ and $\tilde{x}$.

\subsection{Word Importance Ranking}

We rank words by vulnerability scores combining importance and tokenization fragility through two zero-query phases.

\textbf{Attention Importance Scoring (AIS).} Following the leave-one-out attribution framework~\cite{moeller-etal-2024-approximate,liuattribot}, we use a sentence transformer model~\cite{reimers2019sentence} to measure each word's semantic contribution. For word $w_i$, we compute the L2 distance between sentence embeddings before and after removing the word:
\begin{equation}
\label{eq:ais}
\text{AIS}(w_i) = \|f_{\text{st}}(x) - f_{\text{st}}(x_{-w_i})\|_2,
\end{equation}
where $f_{\text{st}}(\cdot)$ is the embedding function of the sentence transformer and $x_{-w_i}$ represents the sentence with word $w_i$ removed.

\textbf{Tokenization Instability Scoring (TIS).} Inspired by the recent discovery that stylistic or non-canonical glyphs can trigger unstable sub-word splits in models~\cite{zhengbroken,sarabamoun2025special}, we measure how stylistic font substitutions affect tokenization fragmentation.
We generate $m$ font substitutions for each word $w_i$ and compute the average fragmentation ratio:
\begin{equation}
\label{eq:tis}
\text{TIS}(w_i) = \frac{1}{m} \sum_{j=1}^{m} \frac{|T_j|}{|T_o|},
\end{equation}
where $T_j$ denotes the tokenization of the $j$-th font substitution, $T_o$ denotes the tokenization of the original word $w_i$, and $|\cdot|$ represents the number of tokens.
The final vulnerability score balances both aspects:
\begin{equation}
\label{eq:vulnerability}
V(w_i) = \alpha \cdot \text{AIS}(w_i) + \beta \cdot \text{TIS}(w_i).
\end{equation}
Words are ranked by descending $V(w_i)$ values.

\input{tables/classification}

\subsection{Attack Mechanism Analysis}

Stylistic font substitutions exploit tokenization differences across model architectures, causing distinct interference patterns.
\ding{182} \textbf{WordPiece tokenization} used by models like DistilBERT converts unrecognized stylistic fonts to \texttt{[UNK]} tokens~\cite{li2020bert}, creating semantic noise through out-of-vocabulary interference.
\ding{183} \textbf{BPE tokenization} in models such as RoBERTa decomposes stylistic fonts into multiple sub-tokens~\cite{radford2019language}, expanding interference compared to single \texttt{[UNK]} replacements.
\ding{184} \textbf{LLM tokenization} exhibits different behavior where models over-interpret stylistic fonts~\cite{yang2024problematic}. When encountering regional indicator symbols, these models activate representations associated with national attributes, creating spurious semantic associations that interfere with text comprehension.
These distinct mechanisms enable SAD to achieve broad effectiveness across different architectures.

\section{Experiment}

\subsection{Experimental Setting}

We mainly evaluate SAD on two common tasks: sentiment classification and machine translation, to demonstrate the broad applicability of our font-based perturbation across different domains.

\textbf{Datasets:} We use SST5~\cite{socher-etal-2013-recursive} and Emotion~\cite{saravia-etal-2018-carer} for sentiment classification, OPUS-100~\cite{zhang-etal-2020-improving} En-Fr and En-Zh for machine translation.

\textbf{Target Models:} We test on DistilBERT~\cite{Sanh2019DistilBERTAD} and RoBERTa \cite{roberta-facebook} for sentiment classification, OPUS-MT~\cite{tiedemann2023democratizing} for machine translation, three LLMs (Qwen2.5-7B~\cite{qwen2.5}, Qwen3-8B~\cite{yang2025qwen3}, Llama3.1-8B~\cite{kassianik2025llama}), and commercial translation APIs (Google~\cite{google_translate}, Baidu~\cite{baidu_translate}, Alibaba~\cite{alibaba_translate}).

\textbf{Baselines:} We compare against methods from TextAttack~\cite{morris2020textattack} including BAE, FD, HotFlip, PSO, TextBugger, Morpheus, Seq2Sick, and additional baselines following the experimental design of Wang et al.~\cite{wang2025multi} such as HQA-Attack~\cite{liu2023hqa}, LimeAttack~\cite{zhu2024limeattack}. Detailed settings are provided in Section~\ref{sec:results}. For fair comparison, all methods operate under comparable experimental conditions.

\textbf{Evaluation Metrics:} Following Wang et al.~\cite{wang2025multi}, we use Attack Success Rate (ASR) measuring successful attacks (↑), Semantic Similarity evaluating meaning preservation between original and adversarial text (↑), and average Query number for efficiency (↓). For translation tasks, we use additional metrics: $\text{RDBLEU} = \frac{(\text{BLEU}(y) - \text{BLEU}(f(x')))}{\text{BLEU}(y)}$ (↑), and $\text{RDchrF} = \frac{(\text{chrF}(y) - \text{chrF}(f(x')))}{\text{chrF}(y)}$ (↑), where $f(x')$ is adversarial translation and $y$ is reference.

\textbf{Implementation} We set $\text{SAD}_{\text{light}}$ with $T=25$ query limit and $\alpha=\beta=0.5$ to balance attack performance and efficiency, using regional indicator symbols, mathematical alphabets and other fonts.

\subsection{Main Results}
\label{sec:results}
\textbf{Traditional Models.}
We evaluate SAD on widely-used pre-trained models for both sentiment classification and machine translation tasks. Tables~\ref{tab:classification} and \ref{tab:opus-100-table} show SAD achieves strong attack performance.

\input{tables/opus-100-mt}

For sentiment classification, SAD$_{\text{light}}$ achieves competitive attack success rates with under 4 average queries while maintaining high semantic similarity above 0.96. SAD$_{\text{strong}}$ reaches over 80\% attack success rates on several settings with single-query efficiency.
For machine translation, SAD consistently outperforms baselines across both tasks. SAD$_{\text{light}}$ maintains high similarity at 0.96 while achieving effective translation degradation. Results confirm font-based perturbations successfully exploit tokenization vulnerabilities across different architectures.

\begin{figure}[ht]
\centering
\includegraphics[width=0.85\columnwidth]{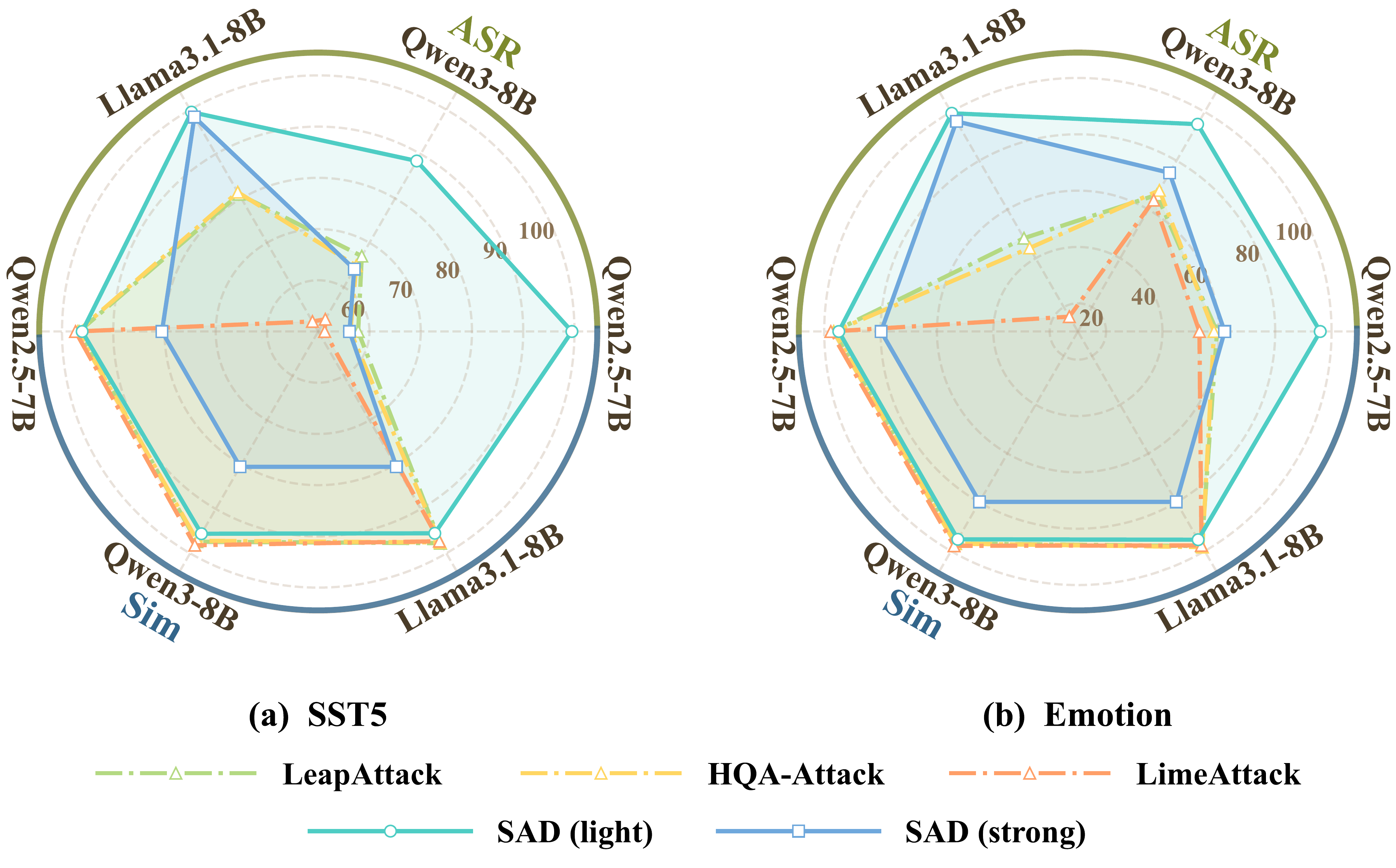}
\caption{Attack performance on large language models.}
\label{fig:llm-radar}
\end{figure}

\textbf{Large Language Models.}
While SAD shows strong performance on traditional models, LLMs have gained widespread adoption in practical applications. We further evaluate SAD on advanced LLMs for security assessment. Following \texttt{Charmer}~\cite{rocamorarevisiting}, we use prompt templates to convert LLMs into classification models. Due to LLMs variability, we only record strict instruction-following samples for rigor. Figure~\ref{fig:llm-radar} shows the results.

SAD$_{\text{light}}$ achieves ASR of 88-99\% with 2.27-7.76 queries, significantly outperforming baselines. We observe that SAD$_{\text{strong}}$ shows weaker performance than SAD$_{\text{light}}$, suggesting extensive stylistic fonts may trigger LLMs alertness to identify original content, while moderate font usage creates subtle interference. For LLMs, font-based perturbations require balancing proportion of stylistic fonts with attack effectiveness.


\input{tables/mt-api}

\textbf{Commercial Applications.}
Furthermore, commercial services are more widely used than research models in practice. We test SAD against Google Translate, Baidu Translate, and Alibaba Translate. Table~\ref{tab:mt-api} presents detailed results.

SAD$_{\text{light}}$ achieves competitive performance with fewer queries, showing query efficiency and high semantic similarity. SAD$_{\text{strong}}$ reveals SAD's attack upper bound with single-query efficiency. The two variants provide different trade-offs between attack scope and query budget, enabling researchers to evaluate model robustness under various threat scenarios. \textit{Results confirm font-based perturbations exploit vulnerabilities in production translation systems.}

\subsection{Discussion}
\begin{figure}[ht]
\centering
\includegraphics[width=0.8\columnwidth]{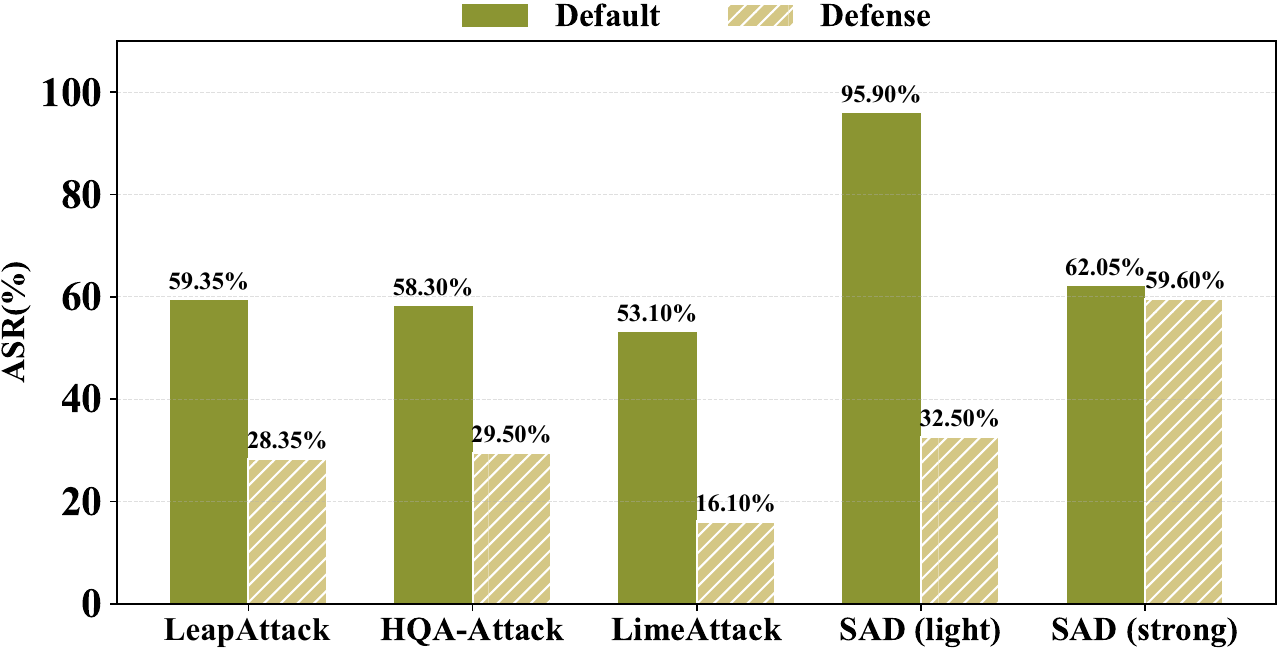}
\caption{Attack performance under defense.}
\label{fig:llm-defense}
\end{figure}

\textbf{Defense Evaluation.}
Modern NLP systems often incorporate defensive preprocessing to counter adversarial attacks. Following prior work, we test SAD against paraphrase defense~\cite{jain2023baseline}. The target model itself reformulates input text while preserving meaning and length before processing.
Due to space constraints, we present defense results on Emotion with Qwen2.5-7B in Figure~\ref{fig:llm-defense}.

Results show that while paraphrase defense effectively reduces all attack methods' performance, SAD consistently outperforms baselines under defensive conditions. This demonstrates SAD's superior robustness, as font-based perturbations are less susceptible to text reformulation compared to traditional adversarial methods.

\begin{figure}[ht]
\centering
\includegraphics[width=0.78\columnwidth]{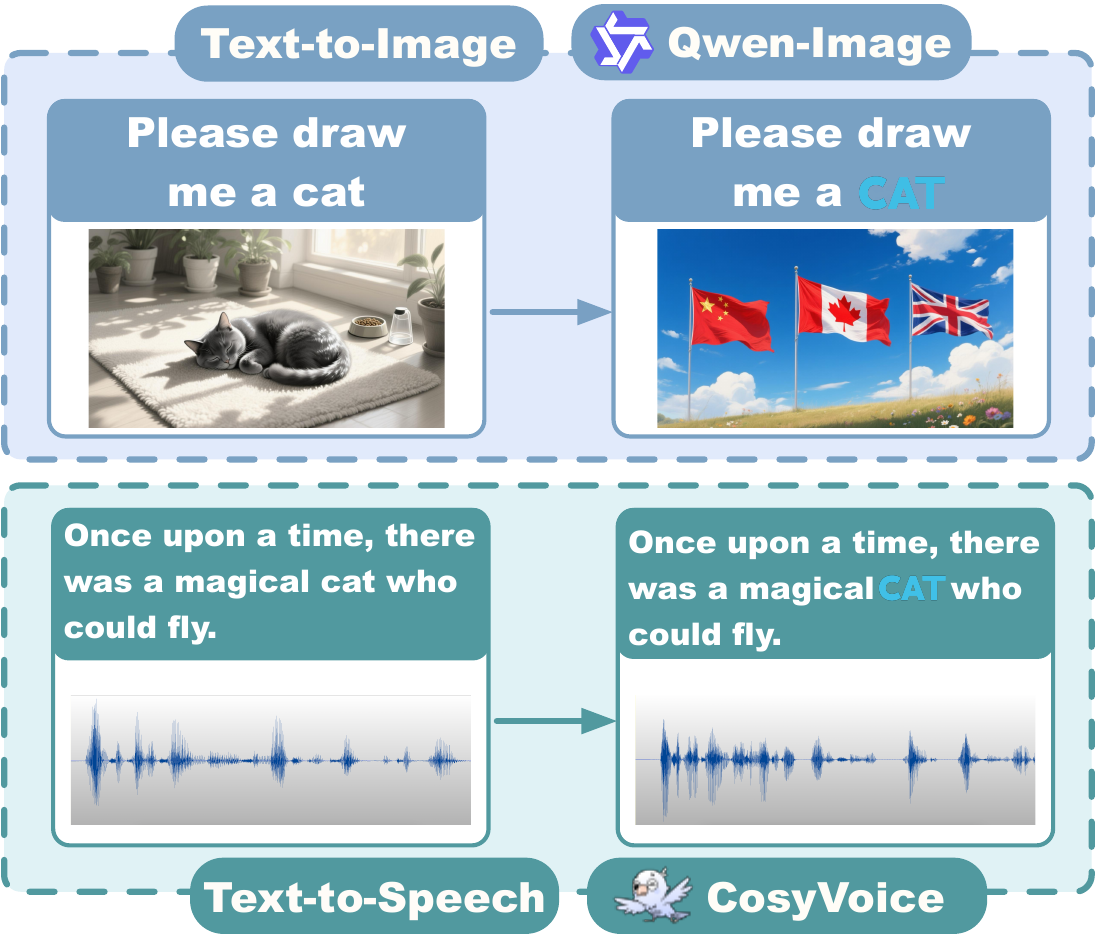}
\caption{For text-to-image task, replacing ``cat" with ``\raisebox{-0.3ex}{\hspace{0.8pt}\includegraphics[height=0.25cm,width=0.62cm]{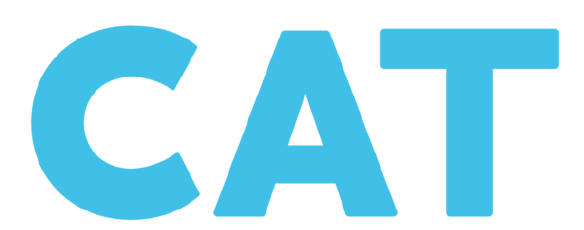}}" causes Qwen-Image to generate flag-related content instead of cats. For text-to-speech task, CosyVoice produces severely distorted audio that even affects surrounding words, making the generated speech difficult to understand.}
\label{fig:extended}
\end{figure}

\textbf{Extended Applications.}
Beyond traditional NLP tasks, SAD's font-based style attack extends to multimodal applications. Figure~\ref{fig:extended} demonstrates SAD's effectiveness on text-to-image and text-to-speech tasks, where stylistic font substitutions trigger unexpected model interference, highlighting the broad applicability of SAD across different tasks and suggesting similar vulnerabilities may exist in other multimodal systems.

\section{Conclusion}

We propose Style Attack Disguise (SAD), exploiting the gap between human and model perception of stylistic fonts. SAD strategically targets words using semantic importance and tokenization instability scoring, then applies font-based perturbations that fool models while preserving readability.
Experiments across traditional models, LLMs, and commercial services show SAD achieves strong attack performance with query efficiency. Results reveal fundamental vulnerabilities when NLP systems process stylistic fonts, raising security concerns. As stylistic fonts continue expanding, style-level attacks pose growing threats to model security. Our future work will focus on developing effective defenses to enhance model robustness against such vulnerabilities, improving overall system security.

\newpage

\bibliographystyle{IEEEtran}
\bibliography{strings}

\end{document}

%% file: tables/classification.tex
\begin{table*}[ht]
\centering
\caption{Attack performance on sentiment classification tasks. For each metric, the best is highlighted in \textbf{bold} and the runner-up is {\ul underlined}.}
\label{tab:classification}
\resizebox{0.9\textwidth}{!}{%
\begin{tabular}{@{}c|cccccc|cccccc@{}}
\toprule
                          & \multicolumn{6}{c|}{SST5}                                                                                                    & \multicolumn{6}{c}{Emotion}                                                                                                  \\ \cmidrule(l){2-13} 
\multirow{-2}{*}{Dataset} & \multicolumn{3}{c|}{DistilBERT}                                                     & \multicolumn{3}{c|}{RoBERTa}           & \multicolumn{3}{c|}{DistilBERT}                                                     & \multicolumn{3}{c}{RoBERTa}            \\ \midrule \midrule
Method                    & ASR(\%)(↑)  & Sim(↑)          & \multicolumn{1}{c|}{Query(↓)}                          & ASR(\%)(↑)  & Sim(↑)          & Query(↓)  & ASR(\%)(↑)  & Sim(↑)          & \multicolumn{1}{c|}{Query(↓)}                          & ASR(\%)(↑)  & Sim(↑)          & Query(↓)  \\ \midrule

BAE                       & 42.71     & 0.888          & \multicolumn{1}{c|}{21.43}                             & 39.14     & 0.887          & 21.48     & 31.55     & 0.925          & \multicolumn{1}{c|}{29.81}                             & 28.50     & 0.924          & 27.97     \\

FD                        & 25.20     & 0.939          & \multicolumn{1}{c|}{12.56}                             & 22.31     & \textbf{0.982} & 9.71      & {\ul47.10}     & 0.948    & \multicolumn{1}{c|}{33.02}                             & 20.75     & 0.979    & 13.36     \\

HotFlip                   & 41.54     & 0.951          & \multicolumn{1}{c|}{11.52}                             & 29.05     & 0.951          & 11.74     & 46.85     & 0.942          & \multicolumn{1}{c|}{10.83}                             & 41.65     & 0.952          & 11.20     \\

PSO                       & 45.16     & 0.954    & \multicolumn{1}{c|}{11.04}                             & 41.49     & 0.954          & 12.38     & 46.05     & 0.945          & \multicolumn{1}{c|}{9.86}                              & 44.95     & 0.964          & 9.88      \\

TextBugger                & 30.36     & \textbf{0.978} & \multicolumn{1}{c|}{31.46}                             & 20.86     & 0.978    & 30.32     & 35.10     & \textbf{0.981} & \multicolumn{1}{c|}{12.61}                             & 29.40     & {\ul0.981} & 12.56     \\

LeapAttack                & 32.58     & 0.953          & \multicolumn{1}{c|}{9.75}                              & 30.09     & 0.944          & 9.54      & 26.30     & 0.934          & \multicolumn{1}{c|}{7.75}                              & 15.50     & 0.939          & 7.66      \\

CT-GAT                    & 29.37     & 0.939          & \multicolumn{1}{c|}{20.92}                             & 24.80     & 0.926          & 37.54     & 25.90     & 0.916          & \multicolumn{1}{c|}{23.67}                             & 26.75     & 0.927          & 23.57     \\

HQA-Attack                & {\ul46.11}     & 0.936          & \multicolumn{1}{c|}{29.35}                             & 39.64     & 0.929          & 29.08     & 37.35     & 0.934          & \multicolumn{1}{c|}{32.86}                             & 35.85     & 0.925          & 23.72     \\

LimeAttack                & 39.10     & {\ul0.975}          & \multicolumn{1}{c|}{29.45}                             & 37.29     & {\ul0.980}         & 29.60     & {36.10}     & {\ul0.980}          & \multicolumn{1}{c|}{29.87}                             & 13.55     & \textbf{0.985}          & 29.83     \\ \midrule
\rowcolor[HTML]{F0E5CE} 

$\text{SAD}_{\text{light}}$      & 44.48  & 0.967                & \multicolumn{1}{c|}{\cellcolor[HTML]{F0E5CE}{\ul3.97}} & {\ul42.13} & 0.967                & {\ul3.96} & 41.20 & 0.960               & \multicolumn{1}{c|}{\cellcolor[HTML]{F0E5CE}{\ul2.47}} & {\ul57.95} & 0.975               & {\ul2.81} \\
\rowcolor[HTML]{F0E5CE} 

$\text{SAD}_{\text{strong}}$     & \textbf{87.10}           & 0.805               & \multicolumn{1}{c|}{\cellcolor[HTML]{F0E5CE}\textbf{1}}         &  \textbf{76.61}         & 0.803               & \textbf{1}         &  \textbf{67.75}         &  0.796              & \multicolumn{1}{c|}{\cellcolor[HTML]{F0E5CE}\textbf{1}}         &  \textbf{82.55}         &  0.801              & \textbf{1}         \\ \bottomrule
\end{tabular}%
}
\end{table*}

%% file: tables/opus-100-mt.tex
\begin{table}[ht]
\centering
\caption{Attack performance on translation tasks. For each metric, the best is highlighted in \textbf{bold} and the runner-up is {\ul underlined}.}
\label{tab:opus-100-table}
\resizebox{0.75\columnwidth}{!}{%
\begin{tabular}{@{}c|c|cccc@{}}
\toprule
                        &                                                      & \multicolumn{4}{c}{OPUS-MT}                                                                                                          \\ \cmidrule(l){3-6} 
\multirow{-2}{*}{Task}  & \multirow{-2}{*}{Method}                             & RDBLEU(↑)                      & RDchrF(↑)                      & Sim(↑)                         & Query(↓)                          \\ \midrule \midrule
                        & PROTES                                               & 0.39                           & {\ul 0.40}                         & 0.74                           & 235.61                            \\
                        & TransFool                                            &0.31                           &0.31                           & 0.87                           & 68.43                            \\
                        & NTA                                                  & 0.39                           & 0.39                           & 0.85                           & 84.34                            \\
                        & Morpheus                                             & 0.26                           & 0.26                           & {\ul0.92}                  & 27.60                       \\
                        & Seq2Sick                                             & 0.29                           & 0.30                           & 0.84                           & 48.94                             \\
                        & kNN                                                  & 0.32                           & 0.32                           & 0.84                           & 60.27                            \\
                        & RA                                                   & 0.26                           & 0.25                           & 0.88                     & 57.66                            \\ \cmidrule(l){2-6} 
                        & \cellcolor[HTML]{F0E5CE}$\text{SAD}_{\text{light}}$  & \cellcolor[HTML]{F0E5CE}{\ul0.55} & \cellcolor[HTML]{F0E5CE}0.39 & \cellcolor[HTML]{F0E5CE}\textbf{0.96} & \cellcolor[HTML]{F0E5CE}{\ul11.53} \\
\multirow{-9}{*}{En-Fr} & \cellcolor[HTML]{F0E5CE}$\text{SAD}_{\text{strong}}$ & \cellcolor[HTML]{F0E5CE}\textbf{0.63}       & \cellcolor[HTML]{F0E5CE}\textbf{0.62}       & \cellcolor[HTML]{F0E5CE}0.85       & \cellcolor[HTML]{F0E5CE}\textbf{1}         \\ \midrule
                        & PROTES                                               & {\ul0.63}                          & 0.60                          & 0.75                           & 172.70                            \\
                        & TransFool                                            & 0.56                           & 0.57                           & 0.86                           & 56.23                            \\
                        & NTA                                                  & 0.61                           & {\ul0.61}                           & 0.84                           & 65.39                            \\
                        & Morpheus                                             & 0.56                           & 0.57                           & {\ul0.91}                  & 24.95                       \\
                        & Seq2Sick                                             & 0.37                           & 0.38                           & 0.87                     & 38.38                             \\
                        & kNN                                                  & 0.48                           & 0.48                           & 0.81                           & 50.63                            \\
                        & RA                                                   & 0.56                           & 0.55                           & 0.82                           & 42.56                             \\
                        \cmidrule(l){2-6}
                        & \cellcolor[HTML]{F0E5CE}$\text{SAD}_{\text{light}}$  & \cellcolor[HTML]{F0E5CE}{\ul0.63}       & \cellcolor[HTML]{F0E5CE}0.55       & \cellcolor[HTML]{F0E5CE}\textbf{0.96}       & \cellcolor[HTML]{F0E5CE}{\ul11.49}          \\
\multirow{-9}{*}{En-Zh} & \cellcolor[HTML]{F0E5CE}$\text{SAD}_{\text{strong}}$ & \cellcolor[HTML]{F0E5CE}\textbf{0.77}       & \cellcolor[HTML]{F0E5CE}\textbf{0.76}       & \cellcolor[HTML]{F0E5CE}0.84       & \cellcolor[HTML]{F0E5CE}\textbf{1}         \\ \bottomrule
\end{tabular}%
}
\end{table}

%% file: tables/mt-api.tex
\begin{table*}[ht]
\centering
\caption{Attack performance on commercial translation services. For each metric, the best is highlighted in \textbf{bold} and the runner-up is {\ul underlined}.}
\label{tab:mt-api}
\resizebox{\textwidth}{!}{%
\begin{tabular}{@{}c|c|cccc|cccc|cccc@{}}
\toprule
                        &                                                      & \multicolumn{4}{c|}{Google Translate}                                                                                                         & \multicolumn{4}{c|}{Alibaba Translate}                                                                                                            & \multicolumn{4}{c}{Baidu Translate}                                                                                                           \\ \cmidrule(l){3-14} 
\multirow{-2}{*}{Task}  & \multirow{-2}{*}{Method}                             & RDBLEU(↑)                         & RDchrF(↑)                         & Sim(↑)                            & Query(↓)                          & RDBLEU(↑)                         & RDchrF(↑)                         & Sim(↑)                            & Query(↓)                          & RDBLEU(↑)                         & RDchrF(↑)                         & Sim(↑)                            & Query(↓)                          \\ \midrule \midrule
                        & PROTES                                               & 0.20                              & 0.19                              & 0.69                              & 45.71                            & 0.38                              & {\ul0.38}                              & 0.68                              & 34.81                          & 0.36                              & {\ul0.37}                              & 0.68                              & 40.75                             \\
                        & TransFool                                            & 0.21                              & 0.21                              & 0.83                              & 18.23                             & 0.24                              & 0.25                              & 0.81                        & 11.27                          & 0.23                              & 0.23                              & 0.81                              & 12.45                             \\
                        & Morpheus                                             & 0.15                              & 0.16                              & {\ul0.89}                     & {\ul 5.63}                       & 0.14                              & 0.15                              & {\ul0.88}                     & {\ul5.10}                   & 0.13                              & 0.13                              & {\ul0.87}                        &  {\ul4.86}                      \\ \cmidrule(l){2-14} 
                        & \cellcolor[HTML]{F0E5CE}$\text{SAD}_{\text{light}}$  & \cellcolor[HTML]{F0E5CE}{\ul0.54}   & \cellcolor[HTML]{F0E5CE}{\ul0.32}    & \cellcolor[HTML]{F0E5CE}\textbf{0.96}     & \cellcolor[HTML]{F0E5CE}10.38 & \cellcolor[HTML]{F0E5CE}{\ul0.61}  & \cellcolor[HTML]{F0E5CE}0.35    & \cellcolor[HTML]{F0E5CE}\textbf{0.96} & \cellcolor[HTML]{F0E5CE}8.15 & \cellcolor[HTML]{F0E5CE}{\ul0.56}   & \cellcolor[HTML]{F0E5CE}{0.33}   & \cellcolor[HTML]{F0E5CE}\textbf{0.96}& \cellcolor[HTML]{F0E5CE}9.29\\
\multirow{-5}{*}{En-Fr} & \cellcolor[HTML]{F0E5CE}$\text{SAD}_{\text{strong}}$ & \cellcolor[HTML]{F0E5CE}\textbf{0.86} & \cellcolor[HTML]{F0E5CE}\textbf{0.91} & \cellcolor[HTML]{F0E5CE}{0.85}          & \cellcolor[HTML]{F0E5CE}\textbf{1}         & \cellcolor[HTML]{F0E5CE}\textbf{0.92} & \cellcolor[HTML]{F0E5CE}\textbf{0.95} & \cellcolor[HTML]{F0E5CE}{0.84}    & \cellcolor[HTML]{F0E5CE}\textbf{1}         & \cellcolor[HTML]{F0E5CE}\textbf{0.86} & \cellcolor[HTML]{F0E5CE}\textbf{0.93} & \cellcolor[HTML]{F0E5CE}{0.85}        & \cellcolor[HTML]{F0E5CE}\textbf{1}         \\ \midrule
                        & PROTES                                               & 0.38                              & 0.37                              & 0.70                              & 57.23                            & {\ul0.64}                              & {\ul0.63}                              & 0.67                              & 39.45                         & {\ul0.68}                              & {\ul0.68}                              & 0.67                              & 46.36                             \\
                        & TransFool                                            & 0.41                              & 0.41                              & {\ul0.84 }                       & 11.53                             & 0.57                              & 0.57                              & 0.82                              & 8.299                          & 0.54                              & 0.56                              & 0.82                              & 7.56                             \\
                        & Morpheus                                             & 0.30                              & 0.40                              & 0.82                              & {\ul5.66}                      & 0.45                              & 0.48                              & 0.83                        & {\ul4.33}                     & 0.42                              & 0.42                              & {\ul0.84}                        & {\ul 4.47}                        \\ \cmidrule(l){2-14} 
                        & \cellcolor[HTML]{F0E5CE}$\text{SAD}_{\text{light}}$  & \cellcolor[HTML]{F0E5CE}{\ul0.43}  & \cellcolor[HTML]{F0E5CE}{\ul0.42}  & \cellcolor[HTML]{F0E5CE}\textbf{0.96} & \cellcolor[HTML]{F0E5CE}15.26 & \cellcolor[HTML]{F0E5CE}{0.53}    & \cellcolor[HTML]{F0E5CE}{0.52}    & \cellcolor[HTML]{F0E5CE}\textbf{0.95} & \cellcolor[HTML]{F0E5CE}12.16 & \cellcolor[HTML]{F0E5CE}{0.50}  & \cellcolor[HTML]{F0E5CE}{0.51}    & \cellcolor[HTML]{F0E5CE}\textbf{0.96} & \cellcolor[HTML]{F0E5CE}12.92 \\
\multirow{-5}{*}{En-Zh} & \cellcolor[HTML]{F0E5CE}$\text{SAD}_{\text{strong}}$ & \cellcolor[HTML]{F0E5CE}\textbf{0.87} & \cellcolor[HTML]{F0E5CE}\textbf{0.89} & \cellcolor[HTML]{F0E5CE}{\ul0.84}       & \cellcolor[HTML]{F0E5CE}\textbf{1}         & \cellcolor[HTML]{F0E5CE}\textbf{0.96} & \cellcolor[HTML]{F0E5CE}\textbf{0.94} & \cellcolor[HTML]{F0E5CE}{\ul0.84}& \cellcolor[HTML]{F0E5CE}\textbf{1}         & \cellcolor[HTML]{F0E5CE}\textbf{0.95} & \cellcolor[HTML]{F0E5CE}\textbf{0.94} & \cellcolor[HTML]{F0E5CE}{0.83}         & \cellcolor[HTML]{F0E5CE}\textbf{1}         \\ \bottomrule
\end{tabular}%
}
\end{table*}